\documentclass[sigconf,anonymous = false]{acmart}

\usepackage{array}
\usepackage[inline]{enumitem}
\usepackage{wrapfig}
\usepackage{booktabs}
\AtBeginDocument{%
  \providecommand\BibTeX{{%
    \normalfont B\kern-0.5em{\scshape i\kern-0.25em b}\kern-0.8em\TeX}}}

\acmYear{2020}\copyrightyear{2020}
\setcopyright{rightsretained}
\acmConference[FAT* '20]{Conference on Fairness, Accountability, and Transparency}{January 27--30, 2020}{Barcelona, Spain}
\acmBooktitle{Conference on Fairness, Accountability, and Transparency (FAT* '20), January 27--30, 2020, Barcelona, Spain}
\acmPrice{}
\acmDOI{10.1145/3351095.3372829}
\acmISBN{978-1-4503-6936-7/20/01}

\begin{document}

\title{Lessons from Archives: Strategies for Collecting Sociocultural Data in Machine Learning}

\author{Eun Seo Jo}
\affiliation{%
 \institution{Stanford University}}
\email{eunseo@stanford.edu}

\author{Timnit Gebru}
\affiliation{%
 \institution{Google}}
\email{tgebru@google.com}
\begin{abstract}
A growing body of work shows that many problems in fairness, accountability, transparency, and ethics in machine learning systems are rooted in decisions surrounding the data collection and annotation process. In spite of its fundamental nature however, data collection remains an overlooked part of the machine learning (ML) pipeline. In this paper, we argue that a new specialization should be formed within ML that is focused on methodologies for data collection and annotation: efforts that require institutional frameworks and procedures. Specifically for sociocultural data, parallels can be drawn from archives and libraries. Archives are the longest standing communal effort to gather human information and archive scholars have already developed the language and procedures to address and discuss many challenges pertaining to data collection such as consent, power, inclusivity, transparency, and ethics \& privacy. We discuss these five key approaches in document collection practices in archives that can inform data collection in sociocultural ML. By showing data collection practices from another field, we encourage ML research to be more cognizant and systematic in data collection and draw from interdisciplinary expertise.
\end{abstract}




\maketitle
\begin{figure}[ht!]
\centering
\includegraphics[width=0.9\columnwidth]{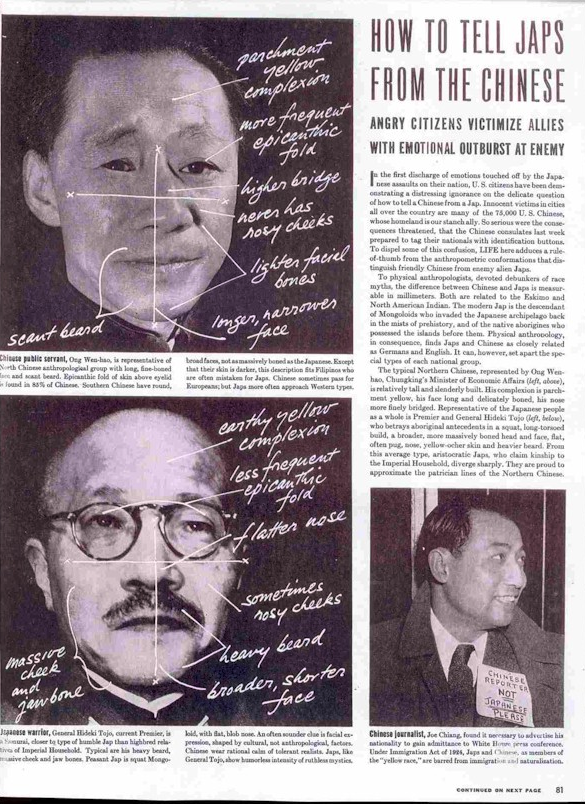} 
\caption{Article from LIFE magazine (Dec. 1941) with two images advising identifiable phenotype differences between Japanese and Chinese ("allies") groups with the intention to spite Japanese Americans following the Japanese bombing of Pearl Harbor.}
\label{lifeMagazine}
\end{figure}

\section{Introduction}
 Data composition often determines the outcomes of machine learning (ML) systems and research. Haphazardly categorizing people in the data used to train ML models can harm vulnerable groups and propagate societal biases. Automated tools such as face recognition software can expose target groups, especially in cases of power imbalance where select institutions have exclusive access to data and powerful models. Historically, biological phenotype traits have been used to single out target groups in moments of public hostility (Fig.~\ref{lifeMagazine}), and similar use cases have been reported today with face recognition technology~\cite{uhugur,perpetual_lineup,japnchin}.\footnote{\textcopyright 1941 The Picture Collection Inc. All rights reserved. Reprinted/Translated from LIFE and published with permission of The Picture Collection Inc. Reproduction in any manner in any language in whole or in part without written permission is prohibited. LIFE and the LIFE logo are registered trademarks of TI Gotham Inc., used under license.} These use cases show the dangers of creating large datasets annotated with people's phenotypic traits.

\begin{table*}
\caption{Lessons from Archives: summaries of approaches in archival and library sciences to some of the most important topics in data collection, and how they can be applied in the machine learning setting. 
}
\smallskip
\centering
\resizebox{\textwidth}{!}{\begin{tabular}{cp{0.8\textwidth}}

\toprule
\centering{Consent} & 

\begin{enumerate*}[itemjoin={\newline}]
\item Institute data gathering outreach programs to actively collect underrepresented data

\item Adopt crowdsourcing models that collect open-ended responses 
    from participants and give them options to denote sensitivity and access
\end{enumerate*}
 \\
\hline
\centering{Inclusivity} &

\begin{enumerate*}[itemjoin={\newline}]
    \item Complement datasets with ``Mission Statements'' that signal commitment to stated concepts/topics/groups
    \item ``Open'' data sets to promote ongoing collection following mission statements  
\end{enumerate*}
\\
\hline
 \centering{Power} &
\begin{enumerate*}[itemjoin={\newline}]
    \item Form data consortia where data centers of various sizes can share resources and the cost burdens of data collection and management
\end{enumerate*}
 \\
 \hline
 \centering{Transparency} &
\begin{enumerate*}[itemjoin={\newline}]
     \item Keep process records of materials added to or selected out of dataset. \item Adopt a multi-layer, multi-person data supervision system.  
\end{enumerate*}
 \\
 \hline
 \centering{Ethics \& Privacy} &
\begin{enumerate*}[itemjoin={\newline}]
    \item Promote data collection as a full-time, professional career. 
    \item Form or integrate existing global/national organizations in instituting standardized codes of ethics/conduct and procedures to review violations
\end{enumerate*}
 \\
 \bottomrule
\end{tabular}}
\label{lessons}
\end{table*}
 On the other hand, in applications such as automated melanoma detection from skin images, it is important to have diverse training data and perform disaggregated testing by various demographic characteristics to ensure that all groups are accurately diagnosed. The quest for large, representative datasets can raise questions of informed consent. Keyes et al. have shown that benchmarks such as those from the National Institute of Standards and Technology in the United States (NIST) consist of data from vulnerable populations taken without consent~\cite{os_nist}. Disaggregated testing also requires gathering potentially sensitive information, and categorizing people into various groups based on demographic information (e.g. gender, age, race, skin type, ethnicity). Many times however, it is unclear how or whether people should be categorized in the first place. While it is important to represent people by their preferred means of representation (e.g. gender identity), other times (such as when documenting instances of discrimination), it may be important to categorize them according to how they are perceived by society. 

Although the manner in which data is gathered, annotated, and used in ML has far reaching consequences, data collection has not been examined with rigor. Holstein et al.'s 2019 summary of critical needs for fair practice among industry ML practitioners identifies the lack of an industry-wide standard for ``fairness-aware data collection'' as an area for improvement across the field. The lack of any systematic process for generating datasets has spurred researchers to call it the ``wild west''~\cite{holstein}.

Recently, an increased focus has been given to data, especially in terms of annotating various demographic characteristics for disaggregated testing, gathering representative data, and providing documentation pertaining to the data gathering and annotation process~\cite{datasheets,modelcards,factsheets}.
However, this move only addresses part of the problem. There are still open questions regarding power imbalance, privacy, and other ethical concerns. As researchers uncover more issues related to ML systems, many have started calling for an interdisciplinary approach to understanding and tackling these issues~\cite{sociotechnical}. Likewise, we call on the ML community to take lessons from other disciplines that have longer histories of addressing similar concerns. In particular, we focus on archives, the oldest human attempt to gather sociocultural data. We outline archives' parallels with data collection efforts in ML and inspiration in the language and institutional approaches for solving these problems in ML. As archives are institutions dealing primarily with documents and photographs, these lessons are best applicable to subfields using unstructured data, such as Natural Language Processing (NLP) and Computer Vision (CV). Of course, archives are just one example of a distant field we can learn from among a wide array of fields. By showing the rigor applied to various aspects of the data collection and annotation process in archives, an industry of its own, we hope to convince the ML community that an interdisciplinary subfield should be formed focused on data gathering, sharing, annotation, ethics monitoring, and record-keeping processes. 

As disciplines primarily concerned with documentation collection and information categorization, archival studies have come across many of the issues related to consent, privacy, power imbalance, and representation among other concerns that the ML community is now starting to discuss. While ML research has been conducted using various benchmarks without questioning the biases in datasets, motives associated with the institutions collecting them, and how these traits shape downstream tasks, archives have
\begin{itemize}
    \item an institutional mission statement that defines the concepts or subgroups to collect data on
    \item full-time curators responsible for weighing the risks and benefits of gathering different types of data and theoretical frameworks for appraising collected data
    \item codes of conduct/ethics and a professional framework for enforcing them
    \item standardized forms of documentation akin to what was proposed in Datasheets for Datasets~\cite{datasheets}
\end{itemize}
In addition, to address issues of representation, inclusivity and power imbalance, archival sciences have promoted various collective efforts such as 

\begin{itemize}
\item community based activism to ensure various cultures are represented in the manner in which they would like to be seen (e.g. Mukurtu\footnote{mukurtu.org})
\item data consortia for sharing data across institutions to reduce cost of labor and infrastructure.
\end{itemize}

We frame our findings about archival strategies into 5 main topics of concern in the fair ML community: consent, inclusivity, power, transparency, and ethics \& privacy. Table~\ref{lessons} summarizes the approaches to these topics in archival studies and how they can be applied to ML. Our results show that archives have institutional and procedural structures in place that regulate data collection, annotation, and preservation that ML can draw from.

The rest of the paper is organized as follows. Section \ref{whatAreArchives} gives an overview of archives and their relevance to ML. Section \ref{supervision} discusses the different levels of supervision in ML and archival data collection. Section \ref{intervention} discusses how data collection can be more ``interventionist''. 
Section \ref{lessonsFromArchives} presents archival approaches to consent, power, inclusivity, transparency and ethics \& privacy and the lessons we can draw from them. 
Section \ref{twoLevels} enumerates how we can implement these approaches at societal and individual levels.
Section \ref{caseStudy} presents a data collection case study to illustrate how these concepts can be applied in practice.
Section \ref{limitations} discusses the limitations of parallels and applications on ML.
Section \ref{conclusion} concludes with open questions and challenges.

\section{What are Archives?}\label{whatAreArchives}
Archives are collections of materials, historical and current, systematically stored for academic, scholarly, heritage, and legacy purposes. As a form of large-scale, collective human record-keeping, archives have existed for thousands of years, long before digital materials. The earliest archives have been state instituted with the purpose of governing the public. The Society of American Archivists (SAA) defines an archive as: An organization that collects the records of individuals, families, or other organizations.\footnote{archivists.org/glossary/terms/a/archives}
Archives may be institutional (eg. United Nations Archives\footnote{search.archives.un.org}), governmental (eg. National Archives and Records Administration\footnote{archives.gov}), foundational (eg. Rockefeller Archive Center\footnote{rockarch.org}), research-oriented (eg. Houston Asian American Archive\footnote{haaa.rice.edu}), among having other objectives. Many modern archives have digital components. In all instances, archives share the objective of collecting human materials as records to be viewed for future uses.

Through years of trials, trends, and debates, archival studies have sophisticated literature on issues of concern in sociocultural material collection. Recent fairness initiatives in the ML community echo procedures and language already developed and used in archival and library communities. To name a few: guidelines for how to label data\footnote{github.com/saa-ts-dacs/dacs} \cite{DACS}; the collection and accessibility of private information \cite{withoutConsent,NaziPrivacy}; sharing datasets across platforms~\cite{libraryConsortia,practicalConsortia}; critical reflections on diversity and inclusivity \cite{archivePower,latino}; theory of appraisal and selection \cite{appraisingManuscripts}. Archivist researchers have developed various schools of data collection; T. R. Schellenberg, F. Gerald Ham, Terry Cook, and Hans Booms have theorized different approaches to appraising documents \cite{recordkeeping}. While the digital component of archiving has been recent, data-aware communities in ML can draw from these historical discussions addressing fundamental questions about using and extracting human information.\\

\section{Differences between Archival and ML Datasets}\label{supervision}
Despite the common goal of collecting data or information, archives and ML datasets differ on a few dimensions. Identifying these differences encourages ML researchers and practitioners to see the possible diversity of data collection practices and equip them with the vocabulary to communicate collection strategies. 

\begin{figure}[t!]
\centering
\includegraphics[width=0.5\textwidth]{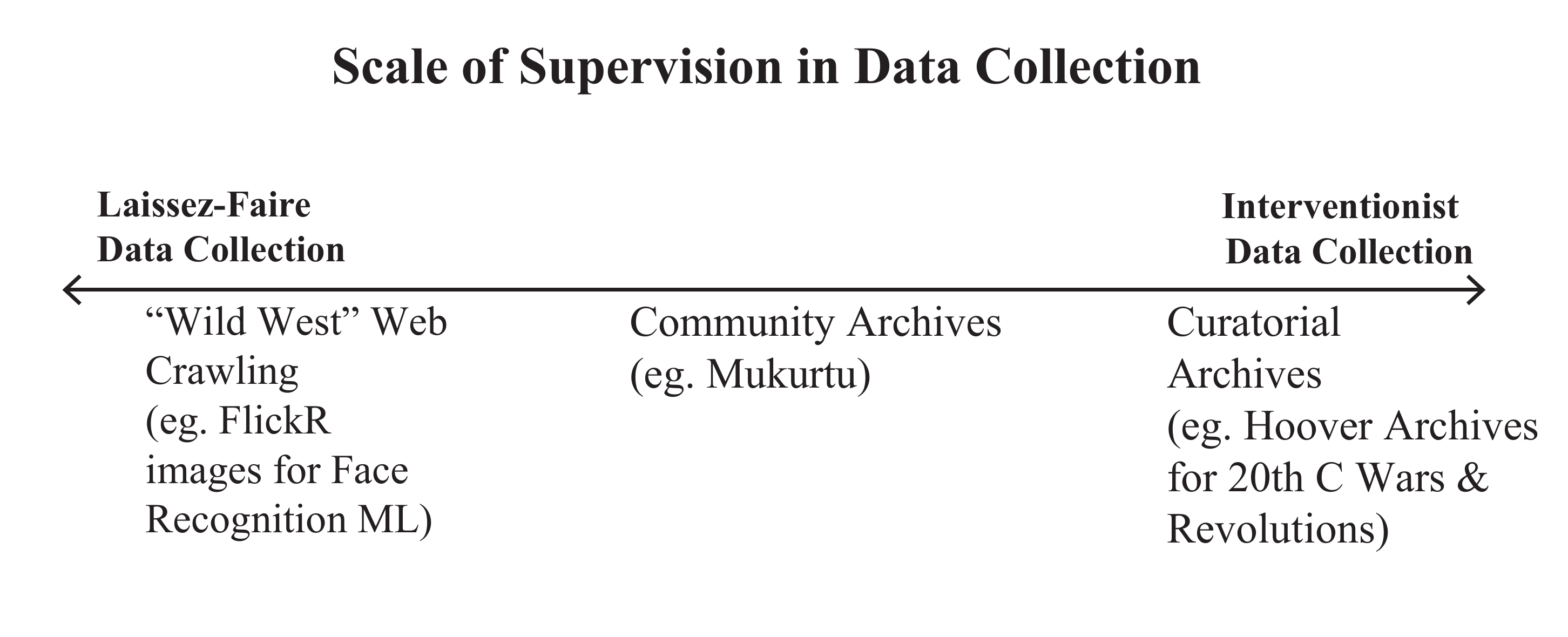} 
\caption{Example categories of data collection practices on supervision scale.}
\label{spectrum}
\vspace{-5mm}
\end{figure}

One area in which current ML data collection practices differ from those of curatorial archives is the level of intervention and supervision. In practice, data collection in significant ML subfields is done without following a rigorous procedure or set of guidelines. While some subfields in ML have fine-grained approaches to data collection, others such as NLP and CV emphasize size and efficiency. This approach often encourages data collection to be indiscriminate \cite{holstein}. Taking data in masses without critiquing its origin, motivation, platform, and potential impact results in minimally supervised data collection. We show the possible spectrum of central supervision in data collection strategies and example archives along the axis in Fig. \ref{spectrum}. No one spot on the spectrum is absolutely preferred in all cases but it is useful to be aware of this measure.

Curatorial archives lie on the other extreme of the intervention spectrum. An archive's raison d'\^etre is defined by the archival Mission Statement (discussed in detail in section \ref{mission}), often a narrow delineation. In selective archives, archivists are trained to evaluate and \textit{filter out} the sources deemed irrelevant or less valuable. Section \ref{appraisal} shows there are several layers of intervention to determine whether given documents or sources are worth adding to the collection. Those deemed not meeting these criteria are removed from the collection. Some archives also target collections from minority groups to diversify their collections as discussed in section \ref{mission}. We call the ``wild west'' method ``laissez-faire'' and the curatorial method a more ``interventionist'' data collection approach. 

ML and archival data collections have differing motivations and objectives. Many ML datasets have the end goal of meeting or beating accuracy measures on defined tasks from training large models \cite{holstein}. Curatorial sets aim to preserve heritage and memory, educate and inform, and have historically tended to be concerned with authenticity, privacy, inclusivity, and rarity of sources \cite{modernArchivesReader}. But these matters of authenticity, privacy, and inclusivity should also inform ML research and defining a project's level of supervision in data collection can help define its objectives.

\section{Need for Interventionist Collection} \label{intervention}
While there may be pros and cons for both laissez-faire and interventionist approaches to data collection, datasets composed without an adequate degree of intervention will replicate biases accrued from multiple levels of filtering. Even before collection, data are subject to two levels of bias - historical and representational \cite{biasLevels}. To minimize these biases, data collection strategies must intervene before applying sampling, weighting, and other balancing techniques. 

Historical bias represents structural, empirical inequities inherent to society that is reflected in the data, such as the historical lack of women presidents in many countries and the underrepresentation of racial minorities in business leadership. When taken wholesale, large indiscriminate ML datasets produce derivatives and outcomes that reflect these biases. For instance, \cite{asianStereo} shows trained word embeddings replicating Asian stereotypes in language data from historical Google Books and the Corpus of Historical American English (COHA).

Representational bias comes from the divergence between the true distribution and digitized input space. This can result from uneven access to digital tools or sociocultural constraints that prevent digitization or preservation. For example, women in the United Arab Emirates are socially stigmatized against photographing their faces, skewing the availability of this type of data to be lower than the true distribution. Some materials may have been destroyed deliberately for political purposes. At the end of World War II, Nazi Gestapo officers burned records of their organization and procedures to avoid prosecution \cite{burnBooks}. These instances bias the availability of certain types of data.

``Natural'' datasets, such as those crawled from the internet, must have an interventionist layer in order to address these inequities at best and at least be used conscientiously. Many ML projects are trained on datasets based on materials found on the internet or already in digital format. Commercial face recognition systems have used FlickR as a source of natural human face images \cite{flickr}. Common sources of natural human language in NLP include crowdsourced material such as Twitter text or data from public platform sites such as Yelp, Wikipedia, IMDB, Stackoverflow, and Reddit. Taken without intervention, these datasets suffer from above sources of biases. For one, materials found on the internet reflect a certain demographic composition. Internet data tend to overrepresent younger generations of users and those from developed countries with internet access \cite{internetYounger,internetDeveloping}. 

But using material from a source that has gone through human supervision is not always ``safe.'' For instance, another common dataset in NLP is text from online news sites (eg. Wall Street Journal, BBC, CNN, Reuters). While published news data go through a screening process supervised by an editorial board, this does not preclude them from having political, topical biases. Fox News tends to write in favor of U.S. political conservatives and the Wall Street Journal tends to produce works on business and commerce relevant topics, generally promoting free trade and other liberal economic policies. The Christian Science Monitor has a religious flavor, catering to its namesake audience. Integral to the news dataset composition is that commercial news providers' sole objective is to keep a healthy readership and market share. They do not produce material aware of its impact on ML models - only the ML researchers using this material will have ownership over that. Critical investigation of the motivations and purpose of the data is an essential component of interventionist data collection.

As discussed in section \ref{lessonsFromArchives}, there are multiple strategies for intervening in data collection to mitigate these levels of bias. We encourage researchers and institutions to be more cognizant and active in data collection.
\begin{table}[t]
\centering

\caption{Parallels in models of archiving and data collection.}\smallskip
\centering
\resizebox{\columnwidth}{!}{
 \begin{tabular}{c  c} 
 \toprule
Archives & ML Data Collection  \\ [0.5ex] 
 \midrule
 Collection Policy \& Mission Statements 
 & Model Objectives, User-centered Needs~\cite{googlepair}\\ 
 Archival Consortia & Data Trusts~\cite{odidatatrusts}  \\
 Participatory \& Community Archives & Crowdsourcing~\cite{projectrespect}\\
 Codes of Conduct \& Ethics & AI Codes of Ethics~\cite{globalAIethics}\\ 
 Appraisal Guidelines \& Records & Datasheets for Datasets~\cite{datasheets}\\ [1ex] 
 \bottomrule
 \end{tabular}
}
\label{parallels}
\vspace{-5mm}
\end{table}

\section{Lessons from Archives}\label{lessonsFromArchives}
We organize issues in fairness accountability transparency and ethics into 5 major abstractions and show how archives have approached them through institutional and interventionist means. Some of these models have parallels in recent initiatives in the ML community as shown in Table \ref{parallels}, allowing us to consider the successes and failures of each approach in the ML context.

\subsection{Inclusivity: Mission Statements \& Collection Policies} \label{mission}
 Inclusivity has become an issue in ML because data collection practices have not been driven by agenda that prioritize fair representation or diversity, rather by tasks or convenience. In ML, many datasets are collected for specific Artificial Intelligence (AI) tasks. For instance, in NLP, the PennTreebank dataset (based on the Wall Street Journal) has served for years as the standard for Part-of-Speech tags \cite{penntree}. Others include OntoNotes for coreference resolution and named entity recognition (NER) and the EuroParl parallel corpus for machine translation (MT) \cite{ontonotes,europarl}. Sometimes, researchers and practitioners build on top of existing datasets motivated by availability and convenience. Based on the titles of accepted papers at the North American Chapter of the Association of Computational Linguistics (NAACL) 2019, at least 7 papers address Twitter, another 5 ``social media'' text, and 9 more work with online news.\footnote{naacl2019.org/program/accepted} This approach produces source-based datasets, where collection methods or questions are defined by the availabiity of datasets.

Setting the data collection agenda by digital availability produces biased data as discussed in section \ref{intervention} and replicates these biases in models. Left without active management of data composition, these methods can lead to limited demographic scope (eg. what kind of users use Reddit?) and render the resulting models to be heavily reliant on the specific source (eg. this model was trained on Reddit data thus reflecting users of Reddit). Researchers have covered the consequences of ML trained on datasets lacking diversity and called for better regulation \cite{predict,gendershades}.

Archives have faced similar critiques of exclusivity. Traditional archives had focused on state and government documents, with the aim to preserve the documents of the governing and social elite. It was not until the rise of social history of the 1960s that archives began recording the lives of the non-elites in earnest \cite{archivists}. Archives address inclusivity by being more effortful and cognizant in defining data collection objectives. One strategy archives abide by is having a Mission Statement. Rather than starting with datasets by availability, data collection in archives starts with a statement of commitment to collecting the cultural remains of certain concepts, topics, or demographic groups. These statements or collection policies could target specializations such as gender minorities in New York or documents on the history of the American West. Many actively announce a commitment to minority collections. All archives have a guiding Mission Statement. Some examples are below:

\begin{quote}
\textbf{Radcliffe College/Schlesinger Library} Schlesinger Library's mission is to document women's lives from the past and present for the future. Its holdings illuminate a vast array of individuals, families, organizations, events, and trends and contain a wealth of resources for the study of social, political, economic, and cultural history. The library's collections are especially rich in the areas of women's rights and feminism, health and sexuality, social reform and activism, work and family life, culinary history, and education and the professions.\footnote{radcliffe.harvard.edu/schlesinger-history-and-holdings}
\end{quote}
\begin{quote}
\textbf{Inland Empire Memories} Inland Empire Memories is an alliance of libraries, archives, and cultural heritage organizations dedicated to identifying, preserving, interpreting, and sharing the rich cultural legacies of diverse communities in Riverside and San Bernardino Counties, a geographical region also known as Inland Southern California. The initiative seeks to increase access to the primary records of individuals and organizations whose work fundamentally shaped the lived experiences of the people in Inland Southern California. A particular emphasis will be placed on those materials that document the lives of peoples and groups underrepresented in the historical record. \footnote{inlandempirememories.org/mission}
\end{quote}
Data collection driven by concepts is more costly and time consuming. It necessitates domain expertise (eg. what are gender minority groups we should be considering?) and exploration (eg. which local organizations should we be contacting?). But a public Mission Statement forces researchers to reckon with their data composition by guiding the data collection process. The search is widened in sources (eg. where can I get images of wedding ceremonies across cultures?) but also filters (eg. is this document relevant to the project's Mission Statement?). The archive community keeps an open dialogue on good-practices for crafting Mission Statements \cite{missionStatements}.


Announcing public Mission Statements also allows datasets to be open to continuous contributions. Researchers can update datasets based on changing sociocultural norms. Coupling datasets with mission statements can inform the research community of their holdings and future amendments to the data. For instance, an image dataset with a mission statement such as \textit{``collecting images of women in various occupations"} can invite the community to continuously contribute data that meet the goals of the statement. 

\subsection{Consent: Community \& Participatory Archives} \label{community}

In ML, crowdsourcing has emerged as a staple approach in collecting human labels for datasets aimed to reduce cost and speed up data collection by outsourcing to human participants. Project Respect crowdsources positive sentiment terms related to the LGBTQ+ community to balance the negative linguistic associations linked with gender minorities online.\footnote{projectrespect.withgoogle.com} However, there are few such examples where ML crowdsourcing relies on open-ended input from communities. Most crowdsourcing mechanisms provide a set of fixed labels for participants to choose from, constraining participants to contribute to the limited options defined by the researchers' agenda. Some crowdsourcing projects imply adversarial relationships between the researcher and the participants, proposing various experimental set-ups featuring competitions and incentive models.\footnote{crowdml.cc/nips2016} 

These methods continue to face criticism for lacking diversity and imposing labels onto individuals. ML researchers without sufficient domain knowledge of minority groups frequently miscategorize data, imposing undesirable or even detrimental labels onto groups. One of the most salient examples is gender. Keyes discusses the socioeconomic perils of gender labeling on trans individuals~\cite{countless}. Hamidi et al. show the overwhelming objection to Automatic Gender Recognition systems among non-binary and non-gender conforming communities \cite{hamidi}. Often, these labels change over time, demonstrating the extent of their social subjectivities. For instance, state-sanctioned race categories have evolved in the United States. In the 1860 U.S. census, one could choose to mark among the options, ``W'' (White), ``B'' (Black), or ``M'' (Mulattos) (U.S. Census Bureau 1860). In 1890, the options expanded to ``White,'' ``Black,'' ``Mulatto,'' ``Quadroon,'' ``Octoroon,'' ``Chinese,'' ``Japanese,'' or ``Indian.'' (U.S. Census Bureau 1890). 

Community archives, also known as tribal archives or participatory archives, are projects of data or document collection in the ownership of the group that is being represented. Projects such as historypin\footnote{historypin.org/en} provide platforms for local communities to define and contribute their own cultural and heritage collections. The aim is to widen the channel of user input in data collection. Community archives are motivated by the need to represent the voices of ``non-elites, the grassroots, the marginalized" \cite{communityArchivesChallenges}. In the anglophone world, community archives have been documenting minority groups since the 1970s ranging from LGBT archives (Hall-Carpenter Archives 1980s) to the Black Cultural Archive established in the U.K. in 1981. Thousands of other self-collecting archives exist today covering various religious, linguistic, class, gender, ethnic, generational, cultural, and regional groups aided by online platforms. In the U.K. alone over a million people are reported to be involved in community archiving~\cite{archivists}. Table \ref{communityArchiveExamples} lists examples of community archives.\footnote{gerberhart.org; lesbianherstoryarchives.org; saha.org.za; saada.org; wcml.org.uk; wisearchive.co.uk}

Community archives serve as an example of how datasets can be opened up for public input, democratize the collection process, and give agency to minority groups to represent themselves. For the purposes of historical archiving, the mission to build an inclusive local heritage and preserve the most complete account possible underlies these initiatives. For instance, women's archives such as the Feminist Archive (1978-), a collection of diaries, personal letters, photographs, among other ephemera contribute to a more complete national history. Thousands of such community archives add diversity to historical records. 

While many of the initiatives have grassroots beginnings, foundations and institutions have been actively funding and promoting archiving in the periphery. For instance, some have aimed to improve participation by distributing equipment ``kits'' to rural regions. These kits include equipment to digitize at-risk audio and video data and equipment to rescue materials from obsolete storage~\cite{kits}. Such programs actively send out resources to local and minority groups to collect their contributions. We have not seen similar initiatives to gather a wider variety of data in ML.

\begin{figure}[t]
\centering
\includegraphics[width=0.9\columnwidth]{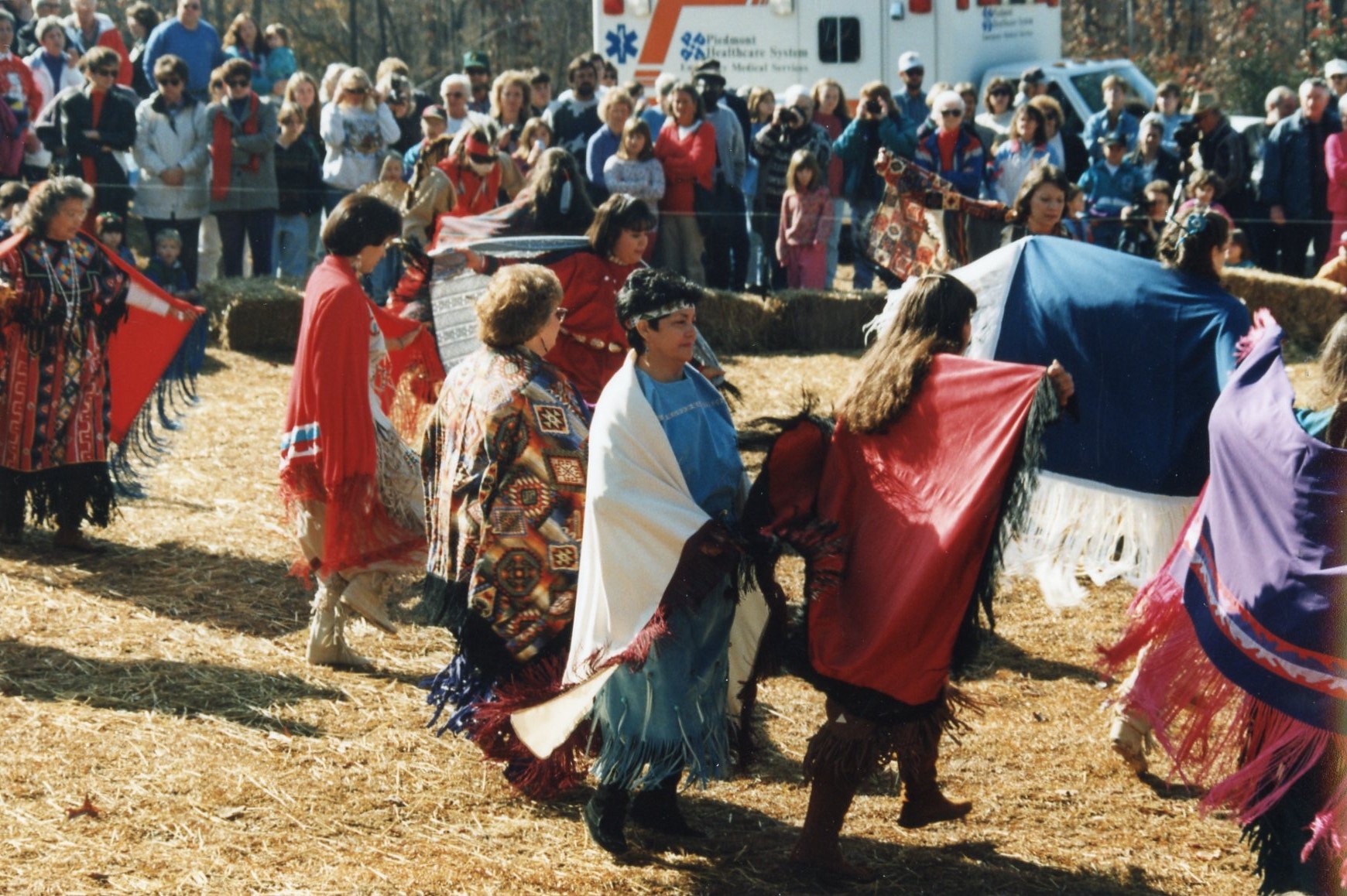} 
\caption{Image of traditional Catawba dance taken in 1993 at the Catawba Indian reservation. \textcopyright Catawba Cultural Preservation Project}
\label{catawba}
\end{figure}

This model of decentralization also enables minority groups to consent to and define their own categorization. Some cultures demand non-Western systems of representation. Mukurtu is an example of a content management system built to allow indigenous communities to house their own materials ``using their own protocols.'' Funded by the National Endowment for the Humanities and the Institute of Museum and Library Services, Mukurtu provides a platform for individuals to upload their data, flag sensitive content, and label their preference for access, use, and circulation all via strictly controlled protocols. These projects give agency to indigenous populations to use their own vocabulary as labels and their choice of images to be integrated into the Mukurtu database. Fig. \ref{catawba} is an example of an image uploaded and labeled by locals~\cite{catawbapic}.

While ML datasets tend to be much larger and homogenous, these examples can serve as starting templates for future community-centered ML data collection procedures. For instance, ML crowdsourcing projects could set up analogous structures around participatory data collection that more actively equip participants with options for access, circulation, and sensitivity. Like Mukurtu, ML datasets that collect international cultural content for instance, can be designed such that participants tag sensitivity and give open-ended input.     

Of course, decentralizing data collection can widen public input but also introduce challenges. Community archives have faced pushback that data quality dilutes as the pool of contributors enlarges~\cite{pliant}. These split viewpoints on the varying degrees of public participation in data collection forewarn ML researchers to thoroughly inspect their own data collection paradigm. Per project, ML researchers must ask, how much supervision, domain expertise, and specialization is needed in collecting data for the scoped project at hand. For instance, an NLP researcher training word embeddings on regional dialects may want to work with anthropologists or other domain experts to reach out to the correct subgroups and accommodate cultural differences.

\newcolumntype{L}{>{\centering\arraybackslash}m{2.6cm}}
\begin{table}
\caption{Example Community Archives}
\centering

\begin{tabular}{LLL} 
 \toprule
Sexuality \& Gender & Political & Social Class\\  
 \midrule
 Gerber/Hart Library and Archives 
 & South African History Archive 
 & Working Class Movement Library\\ 

Lesbian Herstory Archives
& South Asian American Digital Archive
& WISEArchive\\  
 \bottomrule
 \end{tabular}
\label{communityArchiveExamples}
\vspace{-3mm}
\end{table}

\subsection{Power: Data Consortia} \label{consortia}
Implementing systems of ethical data collection demands time, expertise, and resources. Performing disaggregated testing requires the labor cost of annotators for additional demographic labels and handling sensitive data with care requires the resources, expertise and infrastructure to preserve privacy. All these needs disadvantage institutions with low resources. As reported in~\cite{gdpr_stock}, the stocks of smaller startups fell after the announcement of the European General Data Protection Regulation (GDPR) because larger tech companies can leverage more resources to ensure GDPR compliance than can smaller institutions.  

To increase parity in data ownership, archives and libraries have developed a consortial model. In the early twentieth century, groups of archives and libraries set up institutional frameworks and services to share resources and collectively store and distribute holdings called library networks, cooperatives, and consortia \cite{libraryConsortia}. Examples include OCLC, LYRASIS, AMIGOS Library Services, OhioLINK, and MELSA \cite{libraryConsortia}. As of September 2003, the International Coalition of Library Consortia (ICLC) had cited over 170 consortia as members of which over 100 are U.S.-based \cite{consortiaReport}. 

Consortia have several mutual benefits for participating groups. The main advantage is the ability to gain economies of scale. Groups of libraries can make expensive purchases such as subscriptions to academic journals, pool resources for large scale projects such as HathiTrust and DPLA, and reduce technological overhead costs. Also, by communicating holdings with other institutions, consortial members can collectively reduce redundant collections, instead focusing on increasing the size of unique collections. Smaller institutions can also benefit from joining prestigious consortia by increasing visibility to their shared unique holdings. Many libraries participate in multiple consortia: North Carolina State University is a member of nine consortia. Effective consortia benefit both smaller and larger participating institutions by enlarging the size of the consortial collection and making otherwise infeasible projects possible~\cite{libraryConsortia}.

But history has shown that consortia are not without shortcomings. In the 1990s and 2000s, consortia came under criticism for creating bureaucracy, unnecessary committees, delay, and new forms of power imbalance among consortial members. Consortia are funded by membership contributions where members could have varying degrees of financial capacities, prescribing potential power inequities. The additional challenge of consortia in the realm of ML data is the intricate link between profit and data. Many large tech organizations have proprietary datasets they may not share in consortial settings. 

The ML community has been actively discussing consortial arrangements for sharing data. In the U.K., the Open Data Institute's (ODI) data trust initiative has been under development but still at the stage of defining its function and capacity.\footnote{theodi.org} These early models can learn from the trials and errors of library consortia.


\subsection{Transparency: Appraisal Records \& Committee-based Data Collection} \label{appraisal}
The ML fairness community has proposed various measures to address the lack of transparency in data collection and ML model architectures~\cite{datasheets,modelcards,factsheets}. These proposals push for clear communication of the ingredients and procedures that make up ML projects with the public. For example, in Datasheets for Datasets, the authors enumerate questions that elicit researchers to address how the given dataset was collected~\cite{datasheets}. Transparency and accountability form central tenets of archival ethics~\cite{SAACoreEthics,ukCode,ALAethicsCode}. To uphold these principles, archives abide by rigorous record-keeping standards, enhancing not only transparency but also effective operations with other institutions.

Like datasheets, archives have developed detailed standards for data description. Archives use three categories of standards to communicate holdings consistent across institutions: 1) Data content standards specifying the content, order, and syntax of data (ISADG, DACS, RAD), 2) data structure standards, specifying the organization of data (EAD, EAC-CPF), and 3) data value standards, specifying the terms used to describe data (LCSH, AAT, NACO)~\cite{recordkeeping}. Part of the archivist's job is to keep records that adhere to these standards. 

But beyond detailing the contents of data, archives also record the \textit{process} of data collection. Archives are wary that all archival content and records will eventually serve future generations. With this premise, archivists keep records of the decisions and evaluations of the appraisal flow. In rigorous appraisal, the process passes through many layers of supervision by archivists, curators, records creators, and records managers. Table \ref{appraisalExample} shows a multi-level and multi-person example appraisal process.\footnote{Interview with Hoover archivists conducted on 4/10/2019 in Stanford, California, U.S.A.} The levels commensurate with 1 and 2 in the example are committee-based and 3 and 4 are delegated to professional curators and processors.

\begin{table}
\caption{Example Appraisal Flow (Hoover Archives)}\smallskip
  \centering
  \begin{tabular}{p{8cm}}
    \toprule
    \textbf{1. Mission Statement}\\
    \begin{itemize*}[itemjoin={\newline}]
        \item Highest level of agenda formulation determining topics/concepts of concern.
    \end{itemize*}\\
    \textbf{2. Collection Development Policy}\\ 
    \begin{itemize*}[itemjoin={\newline}]
      \item A more specific policy drawn from the Mission Statement about what is collected, what is not, and where and how to search for sources.
    \end{itemize*}\\
    \textbf{3. Appraisal}\\
    \begin{itemize*}[itemjoin={\newline}]
        \item Evaluation based on criteria of whether a given selection of sources is worth collecting.
        \item Asking whether this collection fits the outlines of the mission statement
        \item Evaluating the rarity of the source, the authenticity of its provenance, and its value for future generations.
    \end{itemize*}\\
    \textbf{4. Processing/Indexing (Micro-Appraisal)} \\
    \begin{itemize*}[itemjoin={\newline}]
        \item Processing the sources individually or at the folder/document level, including indexing them and updating the finding aid.
        \item Sources may be discarded out of privacy concerns or for irrelevance.
    \end{itemize*}\\
    \bottomrule
  \end{tabular}
  \label{appraisalExample}
\end{table}

These multiple levels of review and record-keeping are unheard of in ML data collection. While introducing these steps in data collection procedures adds cost and lag in development, these examples from archives serve as models for future strategies in fair ML.

\subsection{Ethics \& Privacy: Codes of Ethics and Conduct} \label{codes}

Researchers have proposed methods and organizations for regulating ethical standards in AI, highlighting the need to monitor privacy and ethical acquisition of data \cite{oxfordAI,dataEthicsCenter}. In the realm of consumer technology, the enforcement of GDPR in 2018 in the European Union (EU) marked a turning point in the history of consumer data protection by instituting non-compliance penalties and sanctions on businesses. No analogous forms of regulation are in place in the United States. Globally, the ISO-IEC JTC 1 established in 1987 and the Institute of Electrical and Electronics Engineers (IEEE) Global initiative on Ethics of Autonomous \& Intelligent Systems are two central standards for AI research and development. But these measures do not address the ethics of data collection let alone provide enforcement mechanisms. Standards, regardless of the importance, become more difficult to implement as objectives move farther from final market transactions and ethical practices in data collection are often overlooked for their distance from the end-product. \cite{oxfordAI}.

 The ethical concerns associated with collecting sociocultural data have a long history in archives. Handling human information exposes archivists to various ethical dilemmas: selecting which documents to toss or keep, granting access to sensitive content, and dealing with intellectual property are just a few of many \cite{intellecturalProperty,foia,culturallySensitive,hardingAffairs}. Several overlapping layers of codes on professional conduct guide and enforce decisions on these matters. Umbrella organizations over archives, libraries, and museums each have individual codes of ethics and conduct. 
The archival codes of ethics via the SAA list the core values of archivists to be to promote access, ensure accountability and transparency, preserve a diverse set of materials, select materials responsibly, and to keep records for the sake of future generations  \cite{SAACoreEthics}. Special collections have specialized protocols to address domain needs such as for preserving rare books or documenting indigenous people \cite{ManuscriptsEthicsCode}.\footnote{atsilirn.aiatsis.gov.au/protocols.php} International groups such as the International Council on Archives (ICA) and the International Council of Museums (ICOM) maintain updated ethical protocols translated into many dozen languages to promote standardized global practices.  

Enforcement of ethics in ML faces challenges due to a lack of an incentive mechanism for researchers and practitioners. While archives may not have found a single foolproof enforcement strategy, there are several features of the archival ecosystem that pressure archivists to comply with ethical guidelines. First, most archivists are full-time professional data collectors. In many organizations, archives work by a membership system whereby breaching the code of conduct could result in losing professional membership~\cite{ukCode}. Many sub-organizations of archival and records collectors have ethics panels or committees that evaluate each alleged violation case by case~\cite{ukCode,ICRMethics}. In ML, promoting full-time employment in data collection will introduce means of raising incentives for data collectors to comply by ethical standards. When a data collector's primary task is selecting and evaluating data under ethics codes and their professional membership is contingent on this task, compliance may be easier to enforce. Because data collection itself is an open ended job, ethics codes can significantly guide the work of a data collector as they do for an archivist. 

Second, establishing cross-institutional organizations that lie above direct employers can help ensure that ethical principles withstand profit-driven motives. Individual members' commitment to the codes of ethics empower data collectors to resist the pressure of their employers to cut corners. While many companies have started to carve out AI Principles\footnote{futureoflife.org/ai-principles}, and organizations such as the Partnership in AI have been formed to create cross-institutional standards\footnote{partnershiponai.org}, they will only be effective if  there are mechanisms by which institutions are held accountable. The ML community can learn from the archives' enforcement strategies.

\section{Two Levels of Actionables}
\label{twoLevels}
We can take action at macro and micro levels to improve ML data collection and annotation. At the macro level, as a community, private institutions, policymakers, and government organizations, we can:
\begin{enumerate}
    \item Congregate and develop data consortia
    \item Establish professional organizations that work by membership to enforce adherence to ethical guidelines
    \item Support community archives 
    \item Develop a subfield dedicated to the data collection and annotation process
\end{enumerate}
At the micro level, as individual researchers, practitioners, and administrators, we can:
\begin{enumerate}
    \item Define and modify Mission Statements
    \item Hire full-time staff on data collection whose performance is also tied to professional standards 
    \item Work towards public datasets
    \item Adopt documentation standards and keep rigorous documentation
    \item Develop more substantial collection development policies with domain expertise and nuance of data source
    \item Make informed committee decisions on discretionary data 
\end{enumerate}
These two levels reinforce and support one another. For instance, it is more effective to administer ethical principles when a central organization outside individual projects holds data collectors accountable.

\section{Case Study: GPT-2 and Reddit}
\label{caseStudy}
To illustrate how lessons from archival history can be applied to ML data collection in practice, we take the example of WebText from OpenAI's GPT-2 language model as a case study~\cite{gpt2}. We choose GPT-2 for our case study due to its timely release, the authors' attention to ethical implications of releasing large language models~\cite{gpt2-modelcards}, and their release of model cards~\cite{gpt2-modelcards} documenting GPT-2's performance and its appropriate use cases warning users of the bias in training data. We explore the motivations for WebText's particular approach and make suggestions for improvements along the 5 dimensions we discussed. While WebText is not publicly available, for the purposes of this case study, we will suppose that it is.

GPT-2 is trained on WebText, a corpus of 8 million documents ($\sim$40G) aggregated by scraping out-links from Reddit pages with at least 3 net upvotes. The \textit{stated} motivation for this data collection process is a performance objective which is to train a language model to be transferred across multiple NLP tasks. In contrast to domain-particular NLP datasets, the focus of WebText was to collect the dataset to be large and diverse, covering a wide variety of contexts, with an emphasis on quality of data. It is assumed to be predominantly English content because the site is Anglophone operating~\cite{gpt2}.

In addition, there are several other \textit{unstated} but inferable motivations to WebText's approach. First, it is cost and time-effective because the data collected is public and online. Second, centering on Reddit increases the likelihood that the dataset is contemporary and comes from a similar distribution as the validation data, which are often also web-scraped datasets. Third, Reddit is a widely enough known platform in the ML community that it is unlikely to require extensive justification of use or explication of content.  

In other words, our inference is that WebText was not optimized for sociocultural inclusivity. The motivation for assembling WebText was to improve the performance of a proposed ML model --hence the focus on diversity of linguistic context and modes-- not to train industry-grade models -- hence the inattention to cultural diversity. 

Consequently from a sociocultural perspective, WebText's composition suggests one hard constraint and a softer limitation. The hard constraint is that all of the content is entirely from documents found on the internet with active links and that are publicly accessible. This limits the source to be written language found online whether that be from uploaded documents or text generated for online use. The softer limitation is that relying entirely on links found on Reddit subjects the dataset to inherit characteristics and biases of Reddit as a platform, such as its user demographic and high turnover rate of content. Pew Internet Research's 2013 survey reveals Reddit users in the United States are more likely to be male, in their late-teens to twenties, and urbanites~\cite{pew}. Thus, the dataset consists of materials of topical relevance to online discussions among this demographic. We classify this web scraping approach as \textit{laissez-faire}, with the only forms of intervention being screening out links if the post had fewer than three net upvotes, and using a blocklist to avoid subreddits containing ``sexually explicit or otherwise offensive content''~\cite{gpt2-modelcards}. Given many studies showing the dubiousness of upvotes as an indication of popularity or quality due to factors such as ``intermediary'' bias, bot votes, and spurious votes, this filtering mechanism could be seen as arbitrary~\cite{Jurgens2017, Glenski, Gilbert2013}. The authors also remove all Wikipedia content to reduce complications with model validation.

We can improve transparency and explicitly delineate the limitations of sociocultural inclusivity by accompanying WebText with a \textbf{Mission Statement} that defines the scope of the material as well as the intention of collection. An example of this is:
\begin{quote}
\textbf{Mission Statement of WebText} WebText is a web-scraped dataset of online documents outlinked from Reddit.com. It currently consists of $\sim$40G of text from 8 million documents. The motivation is to collect 1. large amounts of natural language data 2. across multiple contexts and domains to optimize performance of GPT-2, a language model trained to transfer onto multiple NLP tasks. WebText is composed for this research objective of experimenting with language models, not for commercial use. Other than screening out links with fewer than 3 net upvotes or links to Wikipedia articles, WebText is completely non-interventionist. WebText aims to increase the size of the dataset by continually scraping updated pages on Reddit.  
\end{quote}
Such a Mission Statement outlines what the origin of WebText is, the level of intervention applied in its making, and what applications it is more appropriate for. Furthermore, if WebText were public, part of a \textbf{data consortium}, or accepting external contributions, it could signal how other parties can contribute to this particular dataset.  

However, if WebText, or any other large dataset with the objective of training a language model, had intended to train a comprehensive English language model, a more interventionist approach may be appropriate along with a modified mission statement. An example Mission Statement that targets non-internet American English to complement a Reddit-centric approach is below:
\begin{quote}
\textbf{Complement Dataset Mission Statement} This dataset's mission is to collect language from groups in the United States who do not regularly contribute to internet English. Many NLP models train on English data found on the internet, through large natural text broadcasting sites such as Wikipedia, Reddit, as well as movie and restaurant review sites such as Yelp and IMDB. However, `natural language' English in the United States is spoken by demographics of a wider range than is represented on the internet. This dataset aims to actively collect the variety of English spoken by American culture broadly to contribute to ML systems such as a language model so that they can incorporate expressions, topics of interest, beliefs, and grammatical structures of language from peoples underrepresented on the internet. The dataset is especially focused on colloquial American English across class, education levels, age, and immigration status. 
\end{quote}

A \textbf{collection development policy} can be more systematically developed to address gaps in sociocultural diversity and inclusivity. This is where high to medium level plans are made about how to pursue the Mission Statement and how to collect them. Here we may draw up a list of target demographics whose linguistic data are sparingly digitized or likely to be found outside the relevance of Reddit:

Some basic descriptives of Redditers and the hard constraint of WebText that all its documents are found online give us basic target demographics~\cite{pew}:\\
\begin{itemize*}[itemjoin={\newline}]
    \item Generational Variance: According to PEW statistics, most redditers represent the millennial to post-millennial generations.  An approach to complement WebText would be to identify where language data or topics of relevance to other generations can be found.  
    \item Gender: Redditers are twice as likely to be male as they are female. We can balance the content by examining the genderedness of the links on Reddit.
    \item Internet Access: Residents of areas with limited access to internet are less likely to contribute to Reddit discussions. One strategy to complement is to identify regions of the United States where internet access is limited or usage is low. 
    \item Rural America: Redditers are more likely to be urbanites. This can skew the distribution of topics of relevance. Rural Americans are more likely to have different topical interests and political leanings. For instance, rural Americans are more likely to vote Republican than those in urban sites. 
    \item Immigration: Multi-ethnic/immigrant neighborhoods where English is not the dominant language or residents consume foreign cultures as the dominant source can be another area where non-standard English can be collected. 
\end{itemize*}\\

We can then pursue data collection strategies such as:\\
\begin{enumerate*}[itemjoin={\newline}]
    \item \textbf{Community Archives}: Identify community archives holding materials relevant to demographics of interest. Are there community archives across the country that have already collected language materials?
    \item \textbf{Participatory Schemes}: Generate participatory archives. Send out self-complete kits or staff out to rural communities, multi-ethnic neighborhoods, and areas with low internet usage. Collect oral histories and interviews based on language collection covering a variety of topics.
\end{enumerate*}\\

To ensure ethical supervision in data collection, the project will need to hire \textbf{full-time} staff on data collection and management with membership in extra-archival organizations and draw up or adopt a \textbf{code of ethics} for the project. This ensures the staff involved in data collection are held accountable for ethical violations. WebText is currently nearly completely \textit{non-interventionist}, with no screening for private data. To make considered decisions calculating the gains and risks of adding and subtracting subsets of data, the project can convene \textbf{committees} including domain experts. The committee can intervene on inclusion of sensitive data, extremist data (eg. political extremism), and race/gender targeting data.

Finally, releasing accompanying \textbf{documentation} on WebText's composition can improve the transparency of the content driving the model. Currently there is no explicit mention of any selection process -- if there were any -- of subreddits used, what times of the day or days of the week the data were scraped, obscuring the content. If more interventionist methods are applied, these stages of decision-making can be recorded to further improve transparency.

\section{Limitations}
\label{limitations}
There are several notable caveats to proposals inspired by archival data collection methods. ML datasets tend to be larger and additional studies on scaling these guidelines will inform how transferable they may be in the ML context. In particular, hiring full-time staff, keeping documentation, and implementing collection strategies at large scale incur large overhead, time, and financial costs. Maintaining data consortia at the community level is one way to reduce these costs by economies of scale, resource sharing, and minimizing duplicity. These efforts require substantial coordination and community efforts. But as we have seen large institutions adopt auditing and documentation frameworks such as Datasheets, Model Cards, and Fact Sheets, these investments are not impossible.

ML datasets and archives also have intrinsic differences in motivation. Many ML projects are commercially motivated and corporate funded, driving the focus on internet users and tech consumers and the incentive to keep data proprietary. Archives and libraries are purposed to preserve cultural heritage and diversity. An analysis of business models is a necessary research discussion of its own, beyond the scope of this paper. Further considerations of incentive models in ML can clarify how the community can appropriate these recommendations from archives.  

Finally, interventionist models are not without fault. One concern is that highly selective data collection approaches concentrate power in archivists in determining the portfolio and treating materials ethically. Undue social, political influences to set the agenda are another. However, a multi-layered, and multi-person intervention system still diffuses power among more people and more systematically than when a single ML engineer compiles a dataset. Governance bodies within and across institutions to audit collectors also provide measures to hold collectors accountable. None of these safeguards are currently in place in the ML community.

\section{Conclusion}
\label{conclusion}
This paper has shown how archives and libraries, fields dedicated to human data collection for posterity have grappled with questions of ethics, representation, power, transparency, and consent. These strategies are institutional and procedural, requiring allocated funds and collective efforts of institutions large and small. ML presents additional challenges of addressing a wider audience and fueling commercial products. And while many archives are non-profit and educational, ML datasets are often tied to profit or defense objectives, raising the stakes of problematic data collection. Thus, the investments archives have made in ethical data collection practices may very well be in order in ML.

Archives are not the only place we can learn from. For dealing with direct human subjects, and issues of privacy and representation, we can draw from experimental and field-work driven social sciences such as sociology and psychology. Historians are well-versed in historical context and anthropologists in cultural sensitivities. In navigating an uncharted path, the ML community can look to older fields for examples of successes and failures on comparable matters.


\bibliographystyle{ACM-Reference-Format}
\bibliography{sample-base}

\appendix

\end{document}